\documentclass[sigconf]{acmart}

\usepackage{booktabs} % For formal tables
\usepackage{multirow}
\usepackage{array, multirow}
\usepackage{makecell}
\usepackage{dirtytalk}
\usepackage{xcolor,colortbl}

\setcopyright{rightsretained}
\acmDOI{10.1145/3449639.3459335}

\acmISBN{978-1-4503-8350-9/21/07}

\acmConference[GECCO '21]{the Genetic and Evolutionary Computation Conference 2021}{July 10--14, 2021}{Lille, France}
\acmYear{2021}
\copyrightyear{2021}

\acmPrice{15.00}

\author{Francisco Baeta}
\orcid{0002-9535-2329}
\affiliation{%
  \institution{University of Coimbra, CISUC, DEI}
  \city{Coimbra} 
  \state{Portugal} 
}
\email{fjrbaeta@student.dei.uc.pt}

\author{João Correia}
\orcid{0001-5562-1996}
\affiliation{%
  \institution{University of Coimbra, CISUC, DEI}
  \city{Coimbra} 
  \state{Portugal} 
}
\email{jncor@dei.uc.pt}

\author{Tiago Martins}
\orcid{0003-2638-237X}
\affiliation{%
  \institution{University of Coimbra, CISUC, DEI}
  \city{Coimbra} 
  \state{Portugal} 
}
\email{tiagofm@dei.uc.pt}

\author{Penousal Machado}
\orcid{0002-6308-6484}
\affiliation{%
  \institution{University of Coimbra, CISUC, DEI}
  \city{Coimbra} 
  \state{Portugal} 
}
\email{machado@dei.uc.pt}

\begin{document}

\title{Speed Benchmarking of Genetic Programming Frameworks}

%%% The submitted version for review should be ANONYMOUS
%\author{Francisco Baeta}
%\authornote{Dr.~Trovato insisted his name be first.}
%\affiliation{%
%  \institution{Institute for Clarity in Documentation}
%  \streetaddress{P.O. Box 1212}


\begin{abstract}
Genetic Programming (GP) is known to suffer from the burden of being computationally expensive by design. While, over the years, many techniques have been developed to mitigate this issue, data vectorization, in particular, is arguably still the most attractive strategy due to the parallel nature of GP.
In this work, we employ a series of benchmarks meant to compare both the performance and evolution capabilities of different vectorized and iterative implementation approaches across several existing frameworks. Namely, TensorGP, a novel open-source engine written in Python, is shown to greatly benefit from the TensorFlow library to accelerate the domain evaluation phase in GP.
The presented performance benchmarks demonstrate that the TensorGP engine manages to pull ahead, with relative speedups above two orders of magnitude for problems with a higher number of fitness cases. Additionally, as a consequence of being able to compute larger domains, we argue that TensorGP performance gains aid the discovery of more accurate candidate solutions.

\end{abstract}





\maketitle

\section{Introduction}

Genetic Programming (GP) is a subfield of Evolutionary Computation (EC) that aims to evolve a set of computer programs in a process of continuous stochastic optimization.
The fact that the candidate solutions themselves are represented through code makes GP an automatic problem-solving tool that does not require information about the structure or form of the optimal solution.
On the other hand, GP is historically known for being rather computationally expensive, especially when it comes to the domain evaluation phase~\cite{24}.

However, because the expressions to evaluate remains constant throughout the whole fitness domain, it is possible to vectorize the set of data points, providing an opportunity for parallelism.
In his work, Keijzer \cite{25} studied the benefits of such vectorized approach over the standard case-by-case evaluation for symbolic regressions. As pointed out, such a vectorized method effectively reduces the asymptotic time complexity of evaluating an individual to the number of nodes it contains.

A data vectorization approach is oftentimes coupled with the capabilities of parallel hardware such as Graphics Processing Units (GPUs) \cite{33}.
Whereas a Central Processing Unit (CPU) strives to minimize the latency of operations, a GPU mainly focuses on maximizing data throughput , \textit{i.e.} the amount of information that gets processed in a given time unit.
For GP, a higher data throughput translates to more GP operations per second (GPops) and consequently faster evaluation speeds.

Typically, GPUs operate according to a model called Single Program Multiple Data (SPMD), where many processors simultaneously run the same program/instruction on different inputs. 
This concept applies to domain evaluation in GP, in the sense that in every generation one must run every program in the population on the same set of fitness cases. In recent years, we have witnessed a steady growth in the computing capability of GPUs.
%As a matter of fact, even modern CPUs have seen an increasing concern with overall throughput maximization resulting in higher core counts. While this is true, even top-of-the-line CPUs are still nowhere near the floating-point performance offered by a standard GPU.

%Nevertheless, no matter how fast processors become, one can always make GP setups more computationally taxing, \textit{e.g.} by increasing the number of fitness cases, maximum generations, maximum allowed depth, in the hopes of increasing the accuracy of resulting candidate solutions.

Indeed, even considering that CPUs have seen a throughput increase in recent years, their floating point calculation performance is still nowhere near that of a standard GPU.
In what concerns GP, a throughput oriented architecture is beneficial in the sense that we can always increase the amount of evaluated fitness cases, maximum generations, maximum allowed, etc., in the hopes of improving the fitness of resulting candidate solutions.

Considering how crucial evaluation speeds are for GP applications, in this work we aim to provide a comprehensive study on the performance differences amongst various frameworks using different iterative and parallel data implementations. 
In particular, we analyze TensorGP, a new GP engine written in Python that takes advantage of the TensorFlow library to vectorize the evaluation domain and perform fitness caching \cite{28}. Other frameworks analyzed in this paper include KarooGP, ECJ, TinyGP (Java), DEAP, Evolving Objects (EO), and GPlearn.

\begin{figure*}[h]
  \centering
  \includegraphics[width=6in]{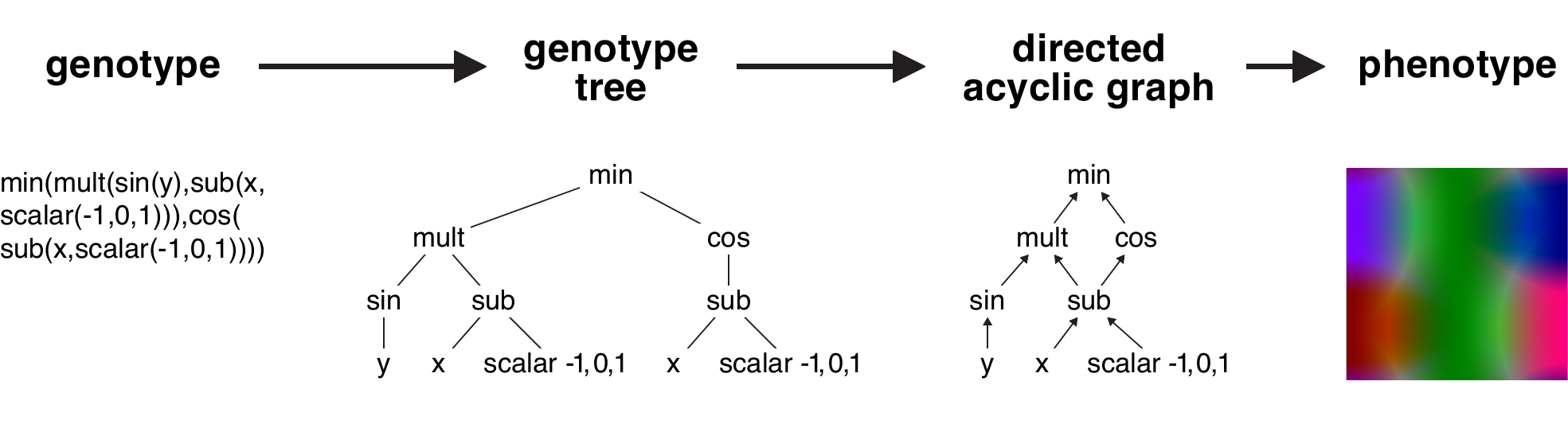}
  \caption{Genotype to phenotype translation phases in TensorGP.}
  \label{fig:teensorgp}
\end{figure*}

% FITNESS MAYDAY
%While our experimental testing mostly focus on running time comparisons, special attention was also given to  the optimization ability of each framework.

The remainder of this paper is structured as follows. Section~2 summarizes the architecture and workflow of TensorGP.
Section~3 introduces the remaining frameworks to test, detailing the experimental setup as well as the problems to benchmark.
Section~4 analyses and discusses gathered results. Finally, Section~5 compiles the main conclusions taken from this study while suggesting several options for future work.

\section{TensorGP}

TensorGP was developed to provide a robust system for seamlessly integrating domain vectorization in GP \cite{28}.
Furthermore, taking advantage of the TensorFlow library, TensorGP manages to heterogeneously distribute computational efforts across different processor types and architectures.

A detailed view of TensorGP's feature set is documented in~\cite{28}.
The following list summarizes the current core features of TensorGP:

\begin{enumerate}
    \item Implementation of standard GP hyperparameters and functionalities: initialization methods, elitism, crossover
    \item Ability to stop and restart the evolutionary process;
    \item Support for custom initial populations in addition to randomly generated ones;
    \item Multiple stop criteria;
    \item Support for customized operators;
\end{enumerate}

\subsection{Domain Evaluation with Tensors}

In TensorGP, the domain of fitness cases to be evaluated is defined by a tensor. Tensors are widely used in many fields of physics and mathematics not only as data structures but also to express multilinear mappings within a vector space \cite{29}. However, in the context of Artificial Intelligence and Computer Science in general, they are often simply defined as a multidimensional array of elements.
By this definition, a tensor is nothing more than a generalization of scalars (\textit{i.e.} a tensor with 0 dimensions), vectors (a tensor with 1 dimension), matrices (a tensor with 2 dimensions), and so on.
Therefore, for a problem domain containing $n$ dimensions, there will be $n$ variables in the terminal set that TensorGP creates to represent its candidate solutions.

The term \say{variables} is misleading in this scenario, as they in fact refer to constant tensors in TensorGP.
In a serial approach, $x$, $y$, $z$, ... will actually be variables because they change according to which element of the domain we are evaluating.
In a vectorized approach, however, we evaluate the entire domain in one vector operation and, as a result, $x$, $y$, $z$, ... will be tensors that contain coordinates within our problem domain.
This means that, for instance, every element in the $x$ tensor will hold the coordinate value of a given fitness case along the $x$-axis.
These tensors are initialized at the beginning of the application and remain unchanged throughout the evolutionary process.

In essence, TensorGP recursively applies tensor operations between these coordinate \say{variables} and other tensors of the terminal set to produce a final output tensor that represents the evaluated program. 
The following subsection details the intermediate representations used to efficiently calculate this output tensor.

\subsection{Representation Overview}

Starting with the traditional way of representing a GP individual as a mathematical expression, there are three translation steps (see Figure \ref{fig:teensorgp}) employed by TensorGP to produce the phenotype for that individual:

\begin{enumerate}
    \item Convert the expression to a Tree Graph.
    \item Transform the Tree Graph into a Directed Acyclic Graph (DAG).
    \item Traverse the obtained graph, recursively performing operations to produce the phenotype.
\end{enumerate}

The resulting phenotype from step 3 is represented as our result tensor that will be used in the next phase of fitness assessment. This resulting phenotype, shown in Figure \ref{fig:teensorgp}, represents a resulting tensor
of rank-3 relating to the height, width and color channels of an image.

It should be noted that step 2 is not implemented by TensorGP itself, but is rather an abstraction provided by TensorFlow to optimize the flow of tensor operations based on its execution mode. TensorFlow has two main modes of execution:
\textit{eager} and \textit{graph}. As opposed to \textit{graph} execution, \textit{eager} mode does not explicitly build the intermediate DAG of operations. 
Although TensorGP was initially developed under the \textit{graph} execution model, it was later upgraded to exploit \textit{eager} execution as this was shown to yield better performance results \cite{28}.

%\subsection{Features}

% nao sei se falar nisto aqui é boa ideia
% como é que eu faço a referencia?

\section{Frameworks}

This section presents the GP frameworks involved in the time comparison benchmarks and the necessary steps taken in the standardization of these systems to achieve commensurable results.

Aside from efficiency and wide-spread use, an essential criterion upon selecting these specific systems lies in the fact that they are free and open-source.

\subsection{Implementations}

Seven different frameworks were tested as shown in Table \ref{tab:frames}.

\begin{table}[hbt!]
\caption{\small Considered GP frameworks along with their respective implementation languages and device support.} \label{tab:frames}
\centering
\begin{tabular}{l|l|l}
\hline
\multicolumn{1}{c|}{\textbf{Framework}} & \multicolumn{1}{c|}{\textbf{Language}} & \multicolumn{1}{c}{\textbf{Processor}}    \tabularnewline \hline
TensorGP     & Python & GPU/CPU \tabularnewline
KarooGP    & Python & GPU/CPU\tabularnewline
ECJ   & Java & CPU\tabularnewline
EO   & C++ & CPU\tabularnewline
TinyGP & Java & CPU\tabularnewline
GPlearn & Python & CPU\tabularnewline
DEAP    & Python & CPU\tabularnewline
\hline
\end{tabular}
\end{table}

We start by detailing TensorGP and KarooGP.
TensorFlow's design philosophy builds upon the principles of heterogeneous computing \cite{1} and as such, TensorGP also inherits this abstraction layer for hardware utilization. For this reason, benchmarks for both CPU and GPU devices will be provided in the next section, regarding this framework. Akin to TensorGP, KarooGP resorts to TensorFlow to vectorize operations over the domain of fitness cases~\cite{2}, which makes it a good candidate for direct comparison with TensorGP running in GPU.
%Moreover, we chose KarooGP for a direct GPU comparison as there already exists several research projects that use this engine \cite{3, 4, 5}.
Moreover, we chose KarooGP as there already exist several research projects that use this engine \cite{3, 4, 5}.

As far as performance is concerned, the main difference between TensorGP and KarooGP resides in the TensorFlow model of operation. Even though both engines represent individuals with a tree-based data structure, KarooGP uses TensorFlow's \textit{graph} execution model, which means that every individual in the population has to be internally compiled into a DAG before the evaluation phase can happen. TensorGP, on the other hand, operates in an \textit{eager} execution model that evaluates the tensor operations imperatively without building intermediate graphs. As a result, \textit{eager} execution aims to eliminate the overhead associated with graph building without sacrificing the benefits provided by graphs~\cite{7}.
In the field of EC in general, the individuals of a population tend to change throughout generations. While many memory and speed optimizations can be achieved by working with a compiled DAG structure, the evolutionary process in GP renders the compilation of an individuals' representation less useful~\cite{28}.

Outside the realm of GPU computing, there are other prominent GP engines worth testing against. In particular, DEAP is a commonly used EC framework that represents the standard for iterative domain evaluation in GP research and literature~\cite{6}. Aside from its popularity, we have chosen to include DEAP as it allows for the quick prototyping of controlled evolutionary environments with little implementation efforts.
Likewise, GPlearn~\cite{15} also capitalizes on the simplification of building GP models by extending the scikit-learn Python machine learning library to perform Symbolic Regression. Additionally, GPlearn supports running the evolutionary process in parallel. However, to the best of our knowledge, the phase that gets parallelized in this framework is the application of genetic operators themselves, not the evaluation of fitness cases. Because the experimental setup used is very evaluation intensive, we did not verify any positive performance gains from running GPlearn with multithreading and therefore this option was disabled.

Another well-established system for evolutionary computation is ECJ~\cite{8}, written in Java. Because most of ECJ's functionality is determined by a hierarchy of parameter files provided by the user, this framework is suited for complex research projects.
Despite ECJ not being primarily designed to run on parallel hardware, its ease of integration enabled Robilliard et al. \cite{10} to develop and compare two GP parallelization schemes running on a NVIDIA GPU.

Similarly to ECJ we must consider EO, an object-oriented framework that provides a flexible set of classes to build EC applications~\cite{9}.
The object-oriented design philosophy is identical to ECJ, with EO providing several classes and interfaces to abstract the evolutionary process. 
%In regards to EO performance, the main difference might just boil down to the programming language itself: C++.
Although the initial learning curve for these last two frameworks might be steeper compared to other GP-based engines such as DEAP, their generalization ability, extensive documentation, and widespread use are the main factors for being included in this comparative performance study.

Lastly, we include TinyGP, a more minimalistic implementation of a GP system that uses a form of tree-based GP. This implementation is based on the Java implementation presented by R. Poli et al.\cite{11}, which in turn is based on the engine originally developed to meet the requirements set out in the TinyGP competition of the Genetic and Evolutionary Computation Conference (GECCO) 2004.

\subsection{Standardization Efforts}

In an effort to level the playing field and achieve comparable results, some modifications were made to the existing frameworks. Hereafter are presented the features that were the object of modifications as well as the reasoning behind them. The source code for all projects used, including modifications, is publicly available. \footnotemark[1]

\footnotetext[1]{Repository available at: https://github.com/AwardOfSky/GP-framework-comparison}
%The source code for all projects used (including modifications) is available at \cite{13}.

GP, like any field of EC, entails stochastic processes related to evolution (\textit{e.g.} primitive and operator selection, population initialization, genetic operators, etc) and as such, every GP framework needs to employ some form of random number generation. Because different implementations call their random generators with different frequencies and at different stages of the evolutionary pipeline, it is not possible to ensure the replication of a given evolutionary path in different engines using the same random seed (even if both initial populations are set to be equal). This means that experimental runs will inevitably follow distinct evolutionary paths and consequently perform a different number of GP operations, which can greatly affect both execution times and the evolution of candidate solutions.

Nevertheless, although a common evolutionary path cannot be set beforehand, we can make efforts to standardize the set of features that the evolutionary process uses, relying upon sufficient experimental runs to average out the inherent randomness.
While it is outside the scope of this work to change existing implementations, because some frameworks do not provide out of the box support for all GP features considered in our experimentation, some implementation efforts were necessary, albeit avoided when possible.

%First and foremost, due to the fact that the benchmark problems considered in
%The biggest implementation struggles were due to the fact that 

%Many of our implementation struggles stemmed from the : we are mainly interested in multivariable problems with a large number of fitness cases (millions in some instances).

Many of the implementation struggles faced stemmed from the nature of our benchmark: we are mainly interested in evolving domains with a large number of fitness cases (millions in some instances).
The amount of data points to be evaluated made the approach of passing a file as input for each individual fitness case unfeasible as such files could easily exceed 1GB for some experiments.
Therefore, engines like KarooGP that followed this design pattern were modified to programmatically evaluate over a linearly spaced domain of data points, which boundaries and granularity can be defined beforehand.
Furthermore, we opted to implement Ephemeral Random Constants \cite{koza1992genetic} across the board, either by generating the values within the tree representation or by having them pre-computed in the terminal set, as TinyGP does. Another implementation concern was the support for GP operations of variable arity, which meant further modifications in some cases.
%the metric implemented across all engines for fitness assessment was the Root Mean Squared Error.

\section{Experimentation}
%4. Experimentation

%In this section, we analyze the performance differences between the GP frameworks previously introduced.
%Two experiments were conducted with results representing an average of 30 runs for every test case.
%The first experiment directly compares the various GP systems in the same controlled environment, measuring not only running time built also fitness optimization performance. 
%The second experiment is aimed at replicating an experimental setup used to solve a real-world problem using the two fastest frameworks derived from the batch of tests in the first experiment. 

In this section, we compare the execution times of various GP systems within a controlled environment, with all results representing an average of 30 runs.
The employed hardware and software is as follows:

%(((Tabela: hardw)))
%CPU - Intel(R) Core(TM) i7-5930K @ 3.50GHz
%GPU - GeForce GTX TITAN X (GM200)
%RAM - 2 \times 16 GB @2.133 Mhz 
%Operative System - Ubuntu 16.04.5 LTS
%Execution environment - Command Line

\begin{table}[hbt!]
\caption{\small Hardware and software specifications used for all experiments.} \label{tab:hard}
\centering
\begin{tabular}{l|l}
\hline
\multicolumn{1}{c|}{\textbf{Component}} & \multicolumn{1}{c}{\textbf{Specification}}    \tabularnewline \hline
CPU     & Intel Core i7-5930K \tabularnewline 
GPU    & NVIDIA GTX TITAN X (12GB)    \tabularnewline
RAM   & 2 $\times$ 16 GB @2.133 Mhz  \tabularnewline 
Operative system   & Ubuntu 16.04.5 LTS  \tabularnewline 
Execution environment & Command line \tabularnewline 
\hline
\end{tabular}
\end{table}

Before proceeding with the experimental results, we must acknowledge that GP is not historically known for having a well-defined set of benchmarking problems and setups.
As a result, some studies analyzed this issue and suggested useful pointers and problem sets aimed at organizing better benchmarking suites in GP \cite{27, 26}. This work intends on putting these guidelines to use in the hopes of achieving a rigorous comparative study.

\subsection{Experimental Setup}
%4.1 old Framework Comparison

All tests employed in this section concern the approximation of the Pagie polynomial function defined by:

 \begin{equation}
 f(x,y)=\frac{1}{1+ x^{-4}} +  \frac{1}{1+ y^{-4}}
  \label{eq:pagie}
\end{equation}

We chose this function as it is considered to be challenging to approximate and is therefore recommended by several GP benchmark articles \cite{32, 27}.

%Besides, many of the studies that recommend problems to use in GP benchmarking also include this polynomial \cite{27, 26}.

As mentioned, the main objective is to compare frameworks directly. In particular, we investigate how increasing the domain size of a problem impacts running times by employing a series of test sets that exponentially increase the number of fitness cases to calculate. Every test set refers to the evaluation of a square two-dimensional domain of values. In the first test set, each framework evaluates a grid of 64 by 64 (4,096 data points).
Each subsequent test set then doubles the length of each side of the grid, effectively quadrupling the total number of evaluations (\textit{i.e.} the second test set has a size of 128 by 128 or 16,384 data points, all the way up to 4,096 by 4.096 or over 16 million points).

%Consequently, the wide range of domain sizes and obtained timing results benchmarked in this experiment warrants the use of a logarithmic scale for graphical representation, as a linear scale would fail to discern between different results.

Because the domain sizes considered in some test sets encompass a wide range of values, our best bet for graphical representation is to use a logarithmic scale as it would be otherwise impossible to distinguish between results. It is also worth noting that because the logarithmic nature of the scale clutters the visualization of the standard deviation values, a table is provided with the corresponding average and standard deviation results.
Aside from domain size, the GP parameterization remains unchanged obeying the following setup:

%(((Tabela: setup1)))
%Runs - 30
%Generations - 50
%Population Size - 50
%Generation method - RHH (population)
%Minimux Initial Deapth - 1
%Minimux Initial Deapth - 10
%Tournament Size - 3
%Mutation Probability - 0.1
%Crossover probability - 0.9
%Fitness metric - RMSE
%Overall allowed max depth - 10
%doimain range - [-5, 5]
%Elitism - 1

\begin{table}[hbt!]
\caption{\small Experimental GP parameters for the framework comparison experiment.} \label{tab:setup}
\centering
\begin{tabular}{l|l}
\hline
\multicolumn{1}{c|}{\textbf{Parameter}} & \multicolumn{1}{c}{\textbf{Value}}   \tabularnewline \hline
Runs     & 30 \tabularnewline
Generations    & 50 \tabularnewline
Population size   & 50 \tabularnewline
Elite size  & 1 \tabularnewline
Tournament size    & 3 \tabularnewline
Mutation probability   & 0.1 \tabularnewline
Mutation probability   & 0.9 \tabularnewline
Minimum initial depth & 1 \tabularnewline
Maximum initial depth & 10 \tabularnewline
Maximum allowed depth & 10 \tabularnewline
Generation method   & RHH (population) \tabularnewline
Fitness metric & RMSE \tabularnewline
Domain range  & [-5, 5] \tabularnewline
\hline
\end{tabular}
\end{table}

As pointed out in Table \ref{tab:setup}, the initial population was generated with the Ramped Half-and-Half (RHH) method.
Typically, there are two ways of implementing this method: one that works over a single tree, generating a root node and creating half of the tree with the grow method using full for the other half, and one that works over the entire population, dividing it into blocks of different depths splitting the number of trees in each block to use either the full or grow method. 
TensorGP uses the second approach to implement RHH. However, this is only but an implementation detail, other frameworks were given the freedom to employ their versions of the RHH algorithm.

Concerning fitness metrics, the Root Mean Squared Error (RMSE) was implemented across all engines.

\begin{figure*}[h]
    \includegraphics[width=4in]{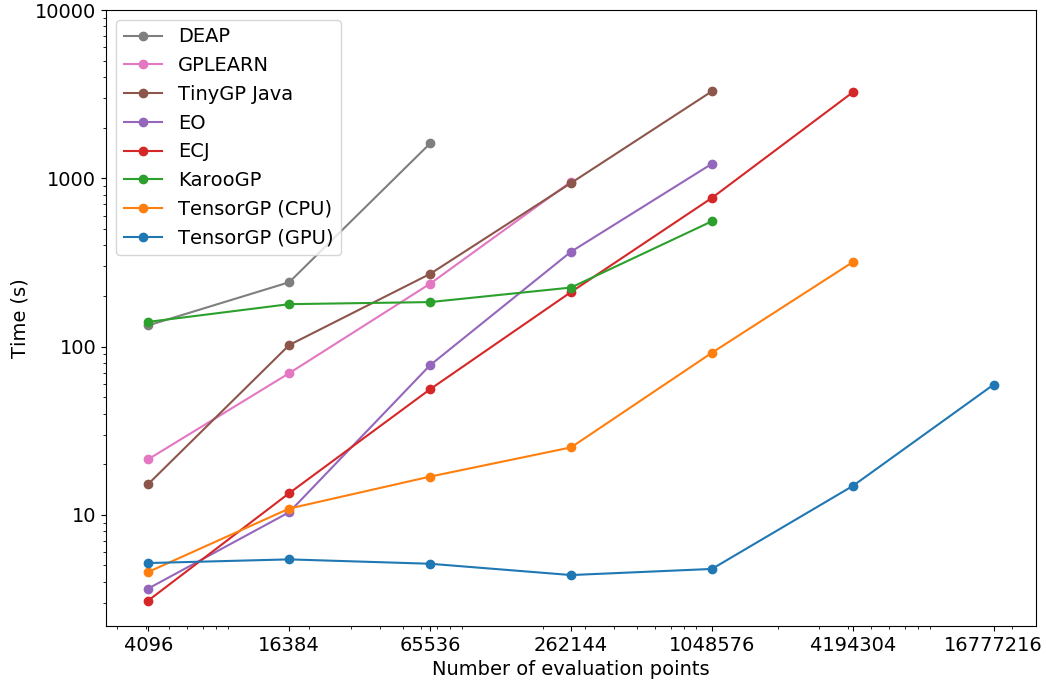}
    \caption{\small 30 run average of total execution time (measured in seconds) across different approaches evaluating a two-dimensional domain ranging from 4,096 data points to over 16 million. Values not shown corresponds to test sets that did not finished either due to memory limitations or by failing to meet the time constraint threshold.}
    \label{fig:im1}
\end{figure*}

No evolutionary methods were re-implemented within the analyzed frameworks to strictly comply with TensorGP's implementation.
For this reason, we are not comparing the evolution of the same set of initial populations using the same genetic operator implementations. However, while it is true that a straightforward comparison can only be achieved over the same evolutionary path, as discussed in the last section, such a comparison is unattainable as it is impossible to reproduce the same evolutionary stochasticity in different frameworks. Besides, implementing the same algorithms across all engines (besides being too daunting of a task) would still not guarantee reproducibility. As a sidenote, GPlearn and KarooGP did not employ elitism. As far as we investigated, elitism is not directly included in the feature set of said frameworks. Still, we assume the possibility of having missed some implementation detail.

Disclaimers aside, we should see this comparison in terms of how much time will it take to execute the same setup across all frameworks considered.

\subsection{Results}

%For the first comparative experiment, a total of 1,320 tests were carried out and are presented here.
%, a total of 1,320 tests were carried out and are presented here.
% isto nao se percebia...
For the performance benchmark, a total of 1,230 tests were carried out. Some approaches that either took too long to finish on average (maximum of 1 hour per experimental run, on average) or crashed due to memory limitations were not considered.
Because TengorGP was benchmarked twice on different hardware, we will refer to each run of the whole array of test sets as relating to an approach rather than a framework to avoid confusion.
Figure \ref{fig:im1} shows the total execution time for all approaches across different domain sizes, each value corresponding to an arithmetic average of 30 runs.

For larger domains, a clear preference towards data vectorization starts to become evident. Besides TensorGP, the only engine to internally employ data vectorization is KarooGP. Even though KarooGP is comparatively slower than most approaches for smaller domains, its running time remains relatively unchanged throughout the first few test sets, as seen in Table \ref{tab:t1}. This phenomenon is easily explained by the internal overhead of building the DAG of GP operations that KarooGP performs with the TensorFlow \textit{graph} execution model. Up until the fourth test set (262,144 points), most of KarooGP execution is spent building the intermediate graphs rather than executing GP operations, hence the jump from last (on the first test set) to third fastest approach.

%\begin{figure}
%\includegraphics[height=1in, width=1in]{fly}
%\caption{A sample black and white graphic
%that has been resized with the \texttt{includegraphics} command.}
%\end{figure}

As demonstrated, the two fastest approaches both used TensorGP, only on different processors.
As the name implies, \say{TensorGP (GPU)} was executed on the GPU described in Table \ref{tab:hard} and \say{Tensor CPU} on the matching CPU.
Due to TensorGP resorting to TensorFlow for domain evaluation, there is always data vectorization being exploited as long as the hardware supports it. We eould be inclined to assume that an approach running on GPU using TensorFlow such as KarooGP would not be slower than another running the same environment under the same library but using a CPU architecture. In reality, aside from the aforementioned graph building overhead specific to KarooGP, we must not underestimate the vectorization capabilities of modern CPUs. Indeed, TensorFlow is able to take advantage of the AVX2 \cite{31} instruction set, available on the CPU used.
Coupled with the absence of graph building overhead, this explains why TensorGP execution on CPU is the second-fastest approach according to our results.

Regarding TensorGP running in GPU, we can identify a clear performance lead throughout all test sets except the first (4,096 data points), where it is outperformed by its CPU counterpart as well as EO and ECJ. However, the lack of relative performance at the start can be easily explained by the fact that, while executing on the GPU, TensorFlow first needs to move the tensor data from the CPU, where it is initialized, to the GPU, and then back again.
Although this memory management overhead limits the performance of TensorGP GPU for smaller domains, the modest performance hit pays off in the remaining test sets. As a matter of fact, by the second test set (16,384 points), TensorGP is already almost twice as fast as the approach running on CPU (see Table \ref{tab:t1}).
TensorGP GPU was the only approach that made feasible the evaluation of the last test set containing over 16 million data points, with an average execution time of just under a minute. Apart from the parallelization power of our GPU, this would not be possible without a sufficiently large VRAM to hold all the tensor data.

\begin{figure*}[h]
    \includegraphics[width=4in]{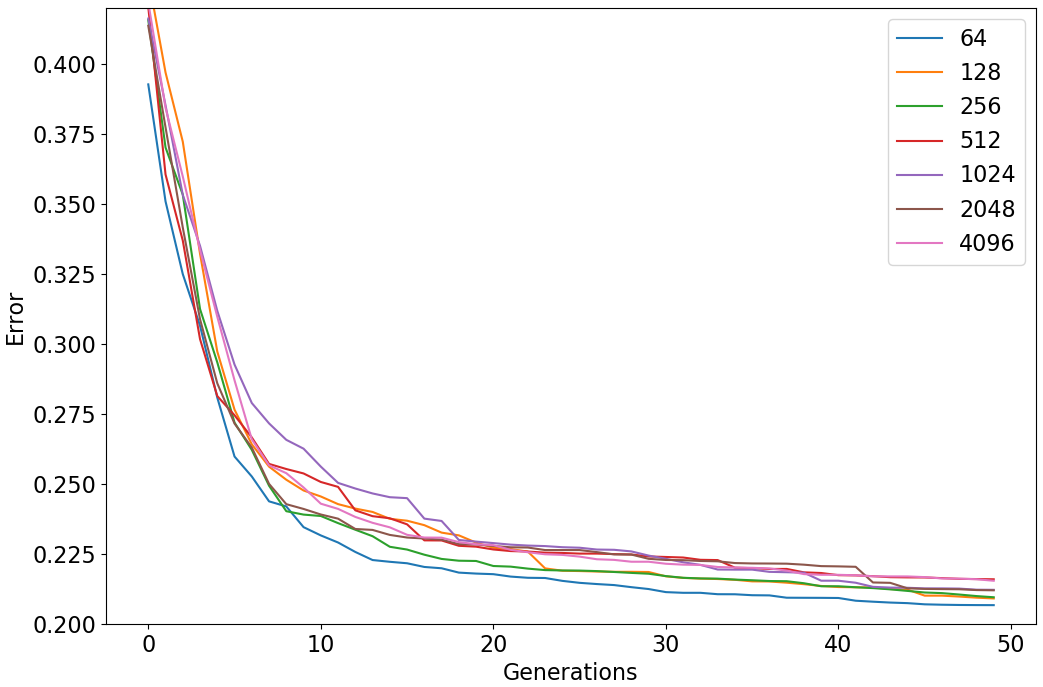}
    \caption{\small 30 run average values regarding the error from the optimal solution of the best individuals across generations for test sets ranging from 4,096 points to over 16 million points. These results correspond to the TensorGP GPU approach, measured with the RMSE fitness metric.}
    \label{fig:imfit}
\end{figure*}

Moreover, it is shown that, for smaller domains, the overhead of moving tensor data is so overwhelming when compared with the time that it takes to execute GP operations, that TensorGP GPU maintains roughly the same overall engine time up to domains with more than 1 million fitness cases (fifth test set). After this point, TensorGP GPU starts to exhibit linear behaviour in proportion to the domain size, as shown by the last two test sets. This happens as we meet the throughput and Video Random Access Memory (VRAM) limitations of the GPU considered, resulting in more data transfers between processors.
TensorGP CPU also shows a similar tendency, with a somewhat sublinear time behaviour up until the fourth test set (262,144 data points), a point from which it increases linearly with an increase in tensor side.
In practice, this phenomenon of near-constant performance followed by a  linear increase in computation time is characteristic of all approaches that use vectorization, especially those with higher parallelization potential such as TensorGP and KarooGP.

As opposed to vectorized approaches, the iterative ones display a linear-like behaviour from the start.
ECJ, written in Java, is a performance winner when it comes to traditional evaluation, only giving its place to EO in the second test set (16,384 points) by a small margin. In turn, EO follows closely behind ECJ, being about 1.5 times slower than this framework for a tensor side of 2,048 (over 4 million data points).
%Aside from being crafted with performance and efficiency in mind, a critical aspect concerning these two frameworks' speed resides in the implementation language itself: Java and C++.
%Still, even comparing a fast iterative system like ECJ to TensorGP runnning on GPU, we observe speedups in excess of 200 times for the execution of the 2.048 test case, further making the case for vectorized approaches.

Akin to ECJ, TinyGP is also implemented in Java but falls short of the evaluation speed achieved by ECJ and EO.
Nevertheless, we must keep in min that TinyGP is mainly an academic exercise meant to show how easily one can set up a GP system, not necessarily a full-blown research tool.
As a consequence, TinyGP has a more minimalistic design that compromises some performance for simplicity.
We also showcase GPlearn, a framework that reveals similar execution times compared to TinyGP even though it is implemented in Python, which is inevitably slower due to the interpretation overheads. For this reason, GPLearn fell short of completing the fifth test case in the allocated time frame, even if by a slight margin.

%Lastly, there is DEAP, only completing the 3 first test cases and thus conming in last considering evaluation speed alone.
Lastly, DEAP is the slowest considering evaluation speed alone, only completing the three first test sets.
Still, in the same manner that TinyGP is not specifically tailored for performance, the main asset associated with DEAP is its ease of use and prototyping rather than speed. In fact, and despite being an EC framework, DEAP was one of the frameworks that required fewer implementation efforts to integrate our experimental setup, the other Python engines that are GP specific.

These tests explore populations with different average depths accentuates the tendency for some values to deviate from the norm, explaining some atypical behaviour in execution times, as demonstrated by the standard deviations in Table \ref{tab:t1}. This was verified in approaches such as DEAP and TensorGP GPU. In DEAP, the 256 test set took considerably longer than expected following a linear behaviour. On the contrary, TensorGP GPU saw a marginal reduction in execution times throughout the first few test sets, mainly due to large data transfer overheads and negligible tensor execution times.

Finally, an analysis of our experimental results would not be complete without looking at performance metrics regarding
fitness evolution.
Figure \ref{fig:imfit} shows the best fitness across generations for all considered test sets performed with TensorGP running on GPU.
Because we perform elitism of size 1, the best-fitted individuals will automatically be promoted to the next generation, ensuring the monotonical decrease of obtained results.
As shown, every test set roughly follows the same behaviour. The first test set displays, on average, slightly better performance across generations, with the largest domain size being marginally worst by the end of the experimental runs.

\begin{table*}[ht]
\caption{\small Execution times, measured in seconds, of 30 run average (AVG) and standard deviation (STD) values across different approaches evaluating a two-dimensional domain which length ranges from 64 to 4,096 fitness cases. Cells corresponding to the fastest approach are highlighted for each test set. Values represented with DNF did not finished either due to memory limitations or time constraints.}
    \centering
\begin{tabular}{ | c | c | r | r | r | r | r | r | r | r | r |p{0.7\linewidth}|}
    \cline{3-10}
 \multicolumn{2}{c|}{} &  \makecell{\textbf{TensorGP} \\ \textbf{(GPU)}} & \makecell{\textbf{TensorGP} \\ \textbf{(CPU)}} & \makecell{\textbf{KarooGP} \\ \textbf{(GPU)}}  & \textbf{ECJ} & \textbf{EO} & \makecell{\textbf{TinyGP} \\ \textbf{(Java)}} & \textbf{GPlearn} & \textbf{DEAP} \\
    \hline
\multirow[t]{2}{*}{}           
            $\mathbf{64 ^ 2}$ & \textbf{AVG} & 5.17 & 4.58 & 140.16 & \cellcolor{blue!25}3.10 & 3.65 & 15.25 & 21.41 & 133.63 \\ %\cline{2-3}
            4,096 & \textbf{STD}  & 2.66 & 2.19 & 47.35 & \cellcolor{blue!25}1.70 & 4.31 & 30.96 & 12.32 & 185.49 \\ \hline
\multirow[t]{2}{*}{}           
            $\mathbf{128 ^ 2}$ & \textbf{AVG} & \cellcolor{blue!25}5.44 & 10.89 & 178.64 & 13.45 & 10.36 & 102.06 & 69.43 & 241.12 \\ %\cline{2-3}
            16,384 & \textbf{STD}  & \cellcolor{blue!25}3.24 & 5.38 & 83.56 & 8.99 & 5.50 & 225.68 & 39.21 & 144.78  \\ \hline
\multirow[t]{2}{*}{}           
            $\mathbf{256 ^ 2}$ & \textbf{AVG} & \cellcolor{blue!25}5.12 & 16.87 & 183.77 & 55.75 & 77.50 & 270.22 & 236.23 & 1610.46  \\ %\cline{2-3}
            65,536 & \textbf{STD}  & \cellcolor{blue!25}1.87 & 8.96 & 75.36 & 40.18 & 79.32 & 407.23 & 84.47 & 3088.09 \\ \hline
\multirow[t]{2}{*}{}           
            $\mathbf{512 ^ 2}$ & \textbf{AVG} & \cellcolor{blue!25}4.39 & 25.16 & 224.22 & 210.77 & 365.95 & 937.06 & 954.02 & DNF \\ %\cline{2-3}
            262,144 & \textbf{STD}  & \cellcolor{blue!25}1.88 & 12.09 & 51.01 & 108.04 & 612.17 & 1513.51 & 545.43 &  DNF\\ \hline
\multirow[t]{2}{*}{}           
            $\mathbf{1,024 ^ 2}$ & \textbf{AVG} & \cellcolor{blue!25}4.77 & 91.85 & 555.33 & 764.34 & 1219.3 & 3287.50 & DNF & DNF \\ %\cline{2-3}
            1,048,576 & \textbf{STD} & \cellcolor{blue!25}2.02 & 52.81 & 167.24 & 430.40 & 1375.67 & 4313.04 & DNF &  DNF\\ \hline
\multirow[t]{2}{*}{}           
            $\mathbf{2,048 ^ 2}$ & \textbf{AVG} & \cellcolor{blue!25}14.85 & 317.78 & DNF & 3244.16 & DNF & DNF & DNF & DNF  \\ %\cline{2-3}
            4,194,304 & \textbf{STD}  & \cellcolor{blue!25}9.82 & 134.72 & DNF & 2058.81 & DNF & DNF &  DNF  &  DNF\\ \hline
\multirow[t]{2}{*}{}           
            $\mathbf{4,096 ^ 2}$ & \textbf{AVG} & \cellcolor{blue!25}59.46 & DNF & DNF & DNF & DNF & DNF & DNF & DNF \\ %\cline{2-3}
            16,777,216 & \textbf{STD}  & \cellcolor{blue!25}24.66 &  DNF  & DNF  &  DNF  &  DNF  &  DNF  &  DNF  &  DNF \\ \hline

\end{tabular}
\\
\footnotesize{\centering DNF stands for \say{Did Not Finish}.}
\label{tab:t1}
\end{table*}

However, a disclaimer must be made about the direct comparison between test sets: because each set evaluates domains of increasing sizes, a lower score on the RMSE fitness metric for a larger domain does not necessarily mean that a worse solution was found.
In fact, more data points condensed in the same domain range translates into less granularity, which in turn allows the assessment against a target that is closer to the optimal solution.
Thereby, the fact that the fitness results are fairly close together only demonstrates that the Pagie Polynomial does not necessarily take advantage of the evaluation over domains with millions of fitness cases (at least not within the considered range).

\section{Conclusion and Future Work}

In this work, we carried out a comparative performance study between an array of GP capable frameworks that are widely used amongst the research community. Additionally, in our study we include TensorGP, a recently developed GP engine written in Python that takes advantage of the TensorFlow library to perform data vectorization.

The experimental results demonstrate that TensorGP achieves faster execution times within the same controlled setup when compared to other systems. Specifically, speedups of up to 100 times were verified when compared to well-established frameworks that perform iterative evaluation such as ECJ or EO as well as a similar vectorization boosted framework (KarooGP).
The aforementioned performance gains are shown to be even more pronounced for the evaluation of domains containing millions of data points and when executing on throughput oriented architectures where GPUs are included. This increase in the number of fitness cases to evaluate provides more information about the optimal solution increasing the potential for approximation and discovery of more accurate candidate solutions.
However, analyzing fitness evolution across generations, we conclude that the benchmarked problem does not significantly benefit from being evaluated on larger domains, as obtained results are predominantly similar throughout all test sets.

More importantly, we must take these results in perspective, realizing that each framework has certain application specializations and that no approach is a perfect fit for all scenarios.
Overall, the fast execution times provided by TensorGP reveal great potential to solve problems with larger domains that require more detailed solutions.

Regarding future work, several endeavors are to be considered. Primarily, the current study can be extended to include other problems, not only pertaining to symbolic regression but also in the areas of classification, predictive modeling, and other metrics.
In particular, the study of problems with a higher pervasiveness of local extrema seems compelling as, theoretically, such problems would reap the benefits furnished by evaluating over high-resolution domains. Likewise, because TensorGP performance makes feasible the exploitation of GP parameters, investigating the influence of individuals with higher depths for certain problems is another possibility to consider.
Furthermore, the inclusion of current real-world GP problems and experimental setups within future testing suites should not be neglected, as these provide irrefutable arguments to the usefulness of any GP framework.

\begin{acks}

This work is funded by national funds through the FCT - Foundation for Science and Technology, I.P., within the scope of the project CISUC - UID/CEC/00326/2020 and by European Social Fund, through the Regional Operational Program Centro 2020 and by the project grant DSAIPA/DS/0022/2018 (GADgET).
We also thank the NVIDIA Corporation for the hardware granted to this research.

\end{acks}

\bibliographystyle{ACM-Reference-Format}
\bibliography{sample-bibliography}

\end{document}